\documentclass{bmvc2k}

\usepackage{boldline,multirow}
\usepackage{booktabs}
\usepackage{xcolor}

\usepackage{forloop}
\newcounter{ct}

\newcommand{\doubleQuote}[1]{\lq\lq{#1}\rq\rq}

\newcommand{\y}{\textbf{y}}

\newcommand{\x}{\textbf{x}}

\newcommand{\T}{\mathsf{T}}

\graphicspath{{figures/}}

\newcommand{\p}{\mathbf{p}}

\newcommand{\br}{\mathbf{r}}

\newcommand{\bv}{\mathbf{v}}

\title{Pan-tilt-zoom SLAM for Sports Videos}

\addauthor{Jikai Lu\textsuperscript{1,}}{lujikai@zju.edu.cn}{2}
\addauthor{Jianhui Chen}{jhchen14@cs.ubc.ca}{1}
\addauthor{James J. Little}{little@cs.ubc.ca}{1}
\addinstitution{
 Department of Computer Science,\\
 University of British Columbia,\\
 Vancouver, Canada
}
\addinstitution{
 College of Computer Science and Technology,\\
 Zhejiang University,\\
 Hangzhou, China
}

\runninghead{Lu, Chen and Little}{Pan-tilt-zoom SLAM for Sports Videos}

\def\eg{\emph{e.g}\bmvaOneDot}

\def\etal{\emph{et al}\bmvaOneDot}

\begin{document}

\maketitle

\footnotetext[1]{{The majority of work was performed when Jikai Lu visited UBC as a Mitacs intern.}}

%that are based on 3D landmarks and 2D-image landmarks
\begin{abstract}
We present an online  SLAM system specifically designed to track pan-tilt-zoom (PTZ) cameras in highly dynamic sports such as basketball and soccer games. In these games, PTZ cameras rotate very fast and players cover large image areas. To overcome these challenges, we propose to use a novel camera model for tracking and to use rays as landmarks in mapping. Rays overcome the missing depth in pure-rotation cameras. We also develop an online pan-tilt forest for mapping and introduce moving objects (players) detection to mitigate negative impacts from foreground objects. We test our method on both synthetic and real datasets. The experimental results show the superior performance of our method over previous methods for online PTZ camera pose estimation. 
\end{abstract}

\section{Introduction}
Visual simultaneous localization and mapping (SLAM) estimates camera poses and environment (\eg landmarks) using stream videos \cite{davison2007monoslam,mur2015orb,bloesch2018codeslam,zhi2019scenecode}, usually assuming that the camera is calibrated and has general movement. While a variety of approaches have been published \cite{concha2016visual,tateno2017cnn,engel2018direct,bresson2017simultaneous, yang2016pop,younes2017keyframe}, very few SLAM methods \cite{lisanti2016continuous} have been explicitly designed for pan-tilt-zoom (PTZ) cameras, which have fixed locations but change pan, tilt angles and zoom levels. These cameras provide detailed views of dynamic scenes and are widely used in sports broadcasting. A robust pan-tilt-zoom SLAM system is beneficial for object tracking and autonomous broadcasting \cite{chen2015mimicking,pidaparthy2019keep,hilton20113d,thomas2017computer}. These real applications motivate our work. 

Our work closely relates to panoramic SLAM systems \cite{civera2009drift,pirchheim2011homography}. These systems assume rotation-only camera motion and track the camera in three degrees of freedom (DoF). Because no parallax is observed, a robust estimation of depth is almost impossible. Researchers have developed many methods to overcome this problem. For example, Pirchheim and Reitmayr \cite{pirchheim2011homography} use plane-induced homographies to represent the cameras between keyframes. Possegger \etal \cite{possegger2012unsupervised} use cylindrical panoramic images to evaluate trajectories of moving objects. Unlike previous methods, we propose to represent the dynamic part of a PTZ camera using pan, tilt angles and focal length. First, this representation has the smallest number of degrees of freedom. As a result, the estimation is more robust than over-parameterized methods when outliers exist. Second, the estimation is practically useful for robotic PTZ cameras which usually have three degrees of control signals (speed of pan, tilt angles and zoom levels).

\begin{figure}[t]
	\centering
	\includegraphics[width=0.95\linewidth]{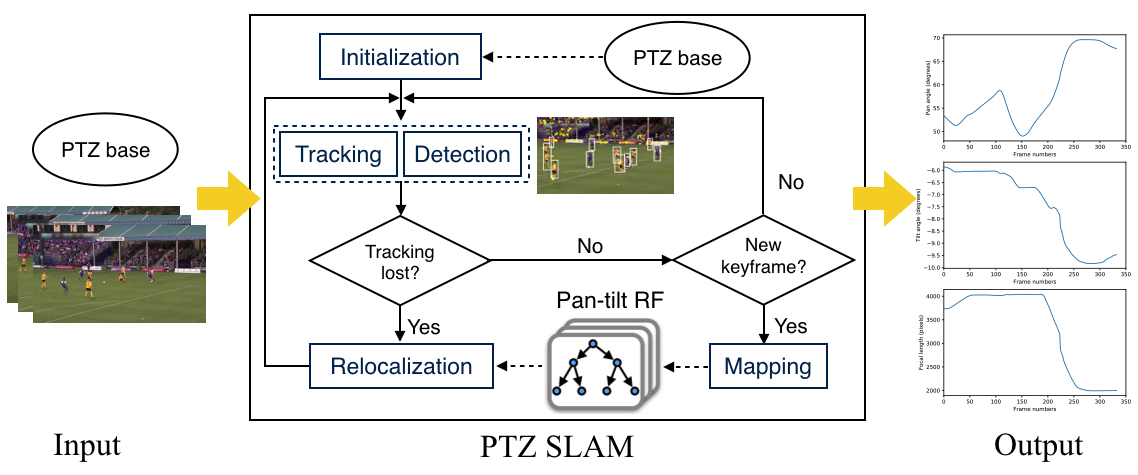}  
    \vspace{-0.1in}
	\caption{System overview. Given a PTZ base and the first camera pose, our system outputs the camera's pan, tilt and focal length for sports videos. Inside the system, we track camera poses and detect players. We also maintain a map by dynamically adding keyframes to a pan-tilt random forest (RF). When the camera tracking is lost, we relocalize the camera using the pan-tilt forest.}
	\label{fig:diagram}
    \vspace{-0.2in}
\end{figure}

Sports videos are challenging for robust camera tracking in long sequences. Sports cameras move very fast so that images can be severely blurred, causing difficulty for feature matching/tracking. Also, moving objects (\eg the players) occlude static objects in the background and inject many outliers for feature tracking. In these situations, sports camera tracking is more challenging than surveillance cameras tracking in which cameras usually do not have very rapid rotations. 

To overcome these challenges, we propose to use a novel camera model \cite{chen2018two} for tracking, extending the two-point method \cite{chen2018two} from a single frame to multiple frames. When camera location and base rotation are known, the two-point method provides the minimum-configuration solution for PTZ cameras. It was proved to be more accurate than homography-based methods. To deal with players, we also integrate player detection results with the SLAM system by discarding keypoints that are on player bodies. 

Figure \ref{fig:diagram} shows the system overview. The system assumes that the camera location and the first-frame camera pose are known. The system outputs a sequence of camera poses (pan, tilt and focal length) from image streams. In the system, the \textit{detection} component detects player bounding boxes using an off-the-shelf object detection method \cite{ren2015faster}. The \textit{tracking} component estimates the optimal solution for pan, tilt and focal length based on the extended Kalman filter (EKF) \cite{davison2007monoslam}. When the tracking is lost, the \textit{relocalization} component recovers the camera pose using an online random forest which is gradually updated by the \textit{mapping} component. 

Our main contributions are: (1) develop a PTZ SLAM system by extending the two-point method \cite{chen2018two} to sequential data and integrating player detection; (2) demonstrate the superior performance of the system on both synthetic and real datasets compared with strong baselines.

\section{Related Work}
\vspace{-0.1in}
\noindent {\bf Image mosaicing and panoramic SLAM:} Image mosaicing estimates the relative pose of the camera when each image is captured \cite{brown2007automatic,lovegrove2010real}. To handle rotation-only camera motion, panoramic SLAMs \cite{gauglitz2012live,pirchheim2013handling} use local panoramic images for mapping. For example, Lisanti \etal \cite{lisanti2016continuous} developed a PTZ camera tracking system. They first build an offline map using multi-scale reference frames. In the online process, the camera parameters are initialized by hardware (camera actuator) and are optimized by 2D image feature matching. They use 2D coordinates in the reference frames as landmarks. Our method and \cite{lisanti2016continuous} are designed for sports and surveillance applications, respectively. Our method is significantly different from \cite{lisanti2016continuous} in two points. First, we build an online map and use one reference frame to initialize the system. Second, we use rays as landmarks for tracking and mapping, which facilitates the EKF SLAM system. 

Combining object detection with SLAM systems is an emerging trend because of the fast development of object detection  \cite{wang2002simultaneous,sun2017improving,zhong2018detect}. For example, Sun \etal \cite{sun2017improving} developed a motion removal approach that filters out moving objects. Zhong \etal \cite{zhong2018detect} developed a robotic vision system that integrates SLAM systems with an object detector to make the two functions mutually beneficial. Our method follows this trend and uses an off-the-shelf object detector \cite{ren2015faster} without training. 

{\bf PTZ camera calibration:} PTZ camera calibration is a fundamental task for surveillance systems and sports applications \cite{thomas2007real,del2010exploiting,rematas2018soccer,chen2019sports}. Previous methods \cite{sinha2006pan,gupta2011using,puwein2011robust} first manually annotate several reference images. Then, they calibrate other images by finding correspondences between a testing image and the reference images. Reference-frame based methods are straightforward but lack of temporal coherence. As a result, they fail when fewer correspondences are available (\eg in blurred images). 

Researchers proposed different camera parameterization methods for PTZ calibration. For example, Hayet \etal \cite{hayet2004robust} first estimate a homography matrix with the reference coordinates. Then, they decompose the pan-tilt-zoom parameters from the estimated homography matrix. The camera model is essential for sequential calibration. Recently, Chen \etal \cite{chen2018two} proposed a novel PTZ camera model. They decompose the projection matrix into two parts. The first part is the camera location and the base (\eg tripod) rotation. This part is fixed and can be estimated by pre-processing. The second part is the pan, tilt and focal length that dynamically change over time. This camera model shows superior performance in single frame calibration. Our work extends \cite{chen2018two} from single image calibration to multiple (sequential) image calibration and tracking.

\section{Method}
We propose a pan-tilt-zoom SLAM system based on the extended Kalman filter and visual feature tracking. The input of our system is the camera location, the tripod base rotation and the camera pose of the first frame. The output is the camera poses of the image sequence. 

\subsection{Preliminaries}
EKF-SLAM \cite{davison2007monoslam} uses a probabilistic feature-based map that represents the current estimates of the camera and all landmarks with uncertainty. The map is represented by a state vector $\hat{\x}$ and a covariance matrix $\mathtt{\Sigma}$. The state vector $\hat{\x}$ is the stacked state estimates of the camera and landmarks $\hat{\x} = [\hat{\x}_v, \hat{\y}_1, \hat{\y}_2, ...]^\T$. The covariance matrix $\mathtt{\Sigma}$ is the uncertainty of these estimates. 

EKF-SLAM has three basic components of \textit{tracking}, \textit{mapping} and \textit{relocalization}. In tracking, the system first uses velocity models to predict camera poses and the projection of landmarks in the image. Then, it tracks point features from the previous frame to the current frame and updates the camera pose and landmarks. In mapping, the system selects frames that have good tracking quality and different camera poses as keyframes. Then, the camera poses and landmarks of keyframes are optimized using bundle adjustment. Finally, the landmarks are memorized as a global map which can be in the form of keyframes, point clouds and neural networks. The relocalization component estimates the camera pose of a lost frame using the map. It either matches the current image against keyframes with known poses, or establishes 2D-to-3D correspondences between keypoints in the current image and landmarks in the scene to estimate camera poses. These three components work together to estimate camera poses robustly.

\subsection{Camera Model and Ray Landmark}
\label{subsec:camera_model}
We use a PTZ camera model from \cite{chen2018two}:
\begin{equation}
    \mathtt{P} = 
    \underbrace{\mathtt{K} \mathtt{Q}_{\phi}\mathtt{Q}_{\theta}}_\text{PTZ} 
    \underbrace{\mathtt{S} [\mathtt{I} | -\mathbf{C}]}_\text{prior},
    \label{equ:ptz}
\end{equation}
\noindent in which $\mathbf{C}$ is the camera location in the world coordinate and $\mathtt{S}$ is the base rotation with three degrees of freedom. $\mathtt{Q}_{\theta}$ and $\mathtt{Q}_{\phi}$ are rotation matrices for pan and tilt angles. $\mathtt{K}$ is the camera matrix in which the focal length $f$ is the only free parameter because we assume square pixels, a principal point at the image center and no lens distortion. 

\begin{figure}
\centering
\includegraphics[width=0.8\linewidth]{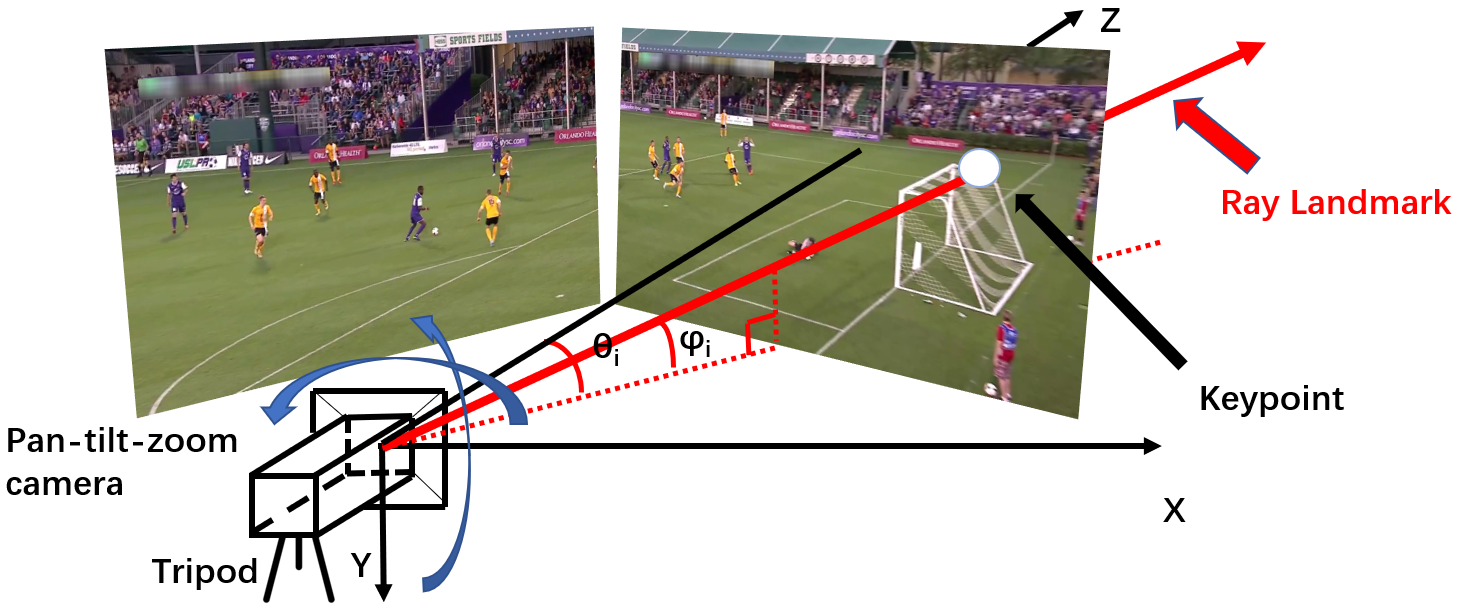}
\vspace{-0.1in}
\caption{The coordinate system and ray landmarks. The camera pose is represented by pan, tilt and focal length in the tripod coordinate. We use a ray (red line) to represent a landmark in the scene. Best viewed in color. }
\label{fig:camera_model}
\end{figure}

There are three coordinates involved with this projection matrix: the world coordinate, the tripod coordinate and the image coordinate (see Figure \ref{fig:camera_model}). First, world points are translated by $\mathbf{C}$ and rotated by $\mathtt{S}$ to the tripod coordinate. Because the tripod is fixed in practice, this $prior$ part $\mathtt{S} [\mathtt{I} | -\mathbf{C}]$ can be pre-estimated. Second, tripod points are rotated and projected to the image coordinate by the $PTZ$ part $\mathtt{K} \mathtt{Q}_{\phi}\mathtt{Q}_{\theta}$. In sports cameras, this part changes over time and is the focus of this work. Starting here, we refer to the pan, tilt angles and focal length as \textit{camera pose}.

In PTZ-SLAM, the camera state $\hat{\x}_v$ comprises pan, tilt and focal length and their velocities $\hat{\x}_v = [\theta, \phi, f, \delta_{\theta}, \delta_{\phi}, \delta_{f}]^\T$. A landmark state $\br_i = [\theta_i, \phi_i]^\T$ is a ray in tripod coordinates. The projection of a ray $\br_i$ to an image is:
\begin{equation}
\mathtt{P}_{ptz}(\br_i) = \mathtt{K} \mathtt{Q}_{\phi}\mathtt{Q}_{\theta}
\begin{bmatrix}
\tan(\theta_i) \\
-\tan(\phi_i) \ \sqrt[]{\tan^2(\theta_i)+1} \\
1
\label{equ:proj_ptz}
\end{bmatrix},
\end{equation}
in which we represent the ray as a 3D point in the plane of $Z=1$. 

When the camera pose is known, we back-project a pixel location $\p_i=[x_i, y_i]^\T$ to a ray by: 
\begin{equation}
    \left \{
    \begin{aligned}
        \theta_{i} &= \arctan(\frac{X}{Z}) \\
        \phi_{i}   &= \arctan ( -\frac{Y}{ \sqrt[]{X^{2} + Z^{2}} } ) 
    \end{aligned} \right.
\label{coor_trans},
\end{equation}

\noindent where:
\begin{equation}
\begin{bmatrix}
X \\
Y \\
Z
\end{bmatrix}
=  \mathtt{Q}^{-1}_{\theta}\mathtt{Q}^{-1}_{\phi}\mathtt{K}^{-1}
\begin{bmatrix}
x_i \\
y_i \\
1
\end{bmatrix}.
\label{equ:proj}
\end{equation}

\noindent This parameterization has the minimum number of degrees of freedom. We will show that our parameterization outperforms other methods such as homography-based methods in tracking. Moreover, our method can take advantage of multi-frame pixel-to-ray correspondences in the relocalization process and thus is more robust than keyframe-based methods. 

\subsection{Online Pan-tilt Forest}
\label{subsec:rf}
We develop an online pan-tilt forest method as the global map for camera relocalization. Unlike previous use of pan-tilt forests \cite{chen2018two}, our method dynamically updates the forests with new keyframes, greatly speeding up the mapping process. We first introduce pan-tilt forests \cite{chen2018two} in brief. Then, we describe the online learning process of the random forest. 

A pan-tilt forest is a regression model that predicts the landmark by its appearance in the image:
\begin{equation}
\hat{\br} = h (\bv),
\end{equation}
\normalsize
where $\bv$ is the feature vector (\eg SIFT descriptor) that describes the image patch. In training, $\{(\bv, \br)\}$ are paired training examples. In testing, the ray $\hat{\br}$ is predicted by the learned regressor $h(\cdot)$. 

A random forest is an ensemble of randomized decision trees. In a tree, the $n^{th}$ internal node splits examples $S_n$ to sub-trees by maximizing of the information gain: 
\begin{equation}
E(S_n, \pi_n) = \sum_{(\x, \br) \in S_n}(\br - \bar{\br})^2 - \sum_{j \in {L,R}}(\sum_{(\x, \br) \in S_n^j}(\br - \bar{\br})^2),
\end{equation}
where $\bar{\br}$ indicates the mean value of $\br$ for all training samples reaching the node, $\pi_n$ is the learned tree parameters. Note that left and right subsets $S_n^j$ are implicitly conditioned on the parameter $\pi_n$. Here we omit the subscript $\p$ of $\br_\p$ and $\x_\p$ for notation convenience. The optimization encourages examples that have similar ray angles to be in same sub-trees.    

Training terminates when a node reaches a maximum depth of $D$ or contains too few examples. In a leaf node, we save the mean values of samples reaching this leaf node $(\bar{\bv}_{l}, \bar{\br}_{l})$. During testing, a sample traverses a tree from the root node to a leaf node. The leaf outputs the $\bar{\br}_{l}$ as the prediction. As a result, a random forest produces multiple ray candidates for a single pixel location. We keep all candidates and choose the one that minimizes the reprojection error in the camera pose optimization process. 

{\bf Online learning:}
During the tracking, keyframes become available in sequential order. We adapt the online random forest method from \cite{saffari2009line} for the training. We assume a forest $h_t(.)$ is available, for example, by training on previous $t$ keyframes. We develop two ways to add new examples $S_{t+1}$ to the forest: update an existing tree or add a new tree. First, we estimate the ratio of correct predictions on $S_{t+1}$ using $h_t(.)$. The correctness is measured by thresholding the angular errors ($0.1^{\circ}$). When the correctness is higher than a threshold (50\%), we update one randomly selected decision tree in $h_t(.)$ by adding new examples $S_{t+1}$. Otherwise, we add a new tree which is trained on $S_{t+1}$ and a random set of examples from previous keyframes. We find the online training decreases the training time without decreasing the prediction accuracy.     

{\bf Camera pose optimization:}
From the pan-tilt forest, we obtain a set of pixel locations and predicted ray pairs $\{(\p, \hat{\br})\}$. The camera pose optimization is to minimize the re-projection error:
\begin{equation}
\{\theta, \phi, f\} = \arg \min_{\mathtt{P}} \sum_{i}{\|\p_i - \mathtt{P}(\hat{\br}_i) \|^2}.
\end{equation}
Because the predictions from the pan-tilt forest may still have large errors, we use the RANSAC method \cite{fischler1981random} to remove outliers. Moreover, we use the two-point method \cite{chen2018two} to estimate the initial camera pose inside of the RANSAC method.

\subsection{Player Detection}

\begin{figure}[t]
	\centering
	\includegraphics[width=0.85\linewidth]{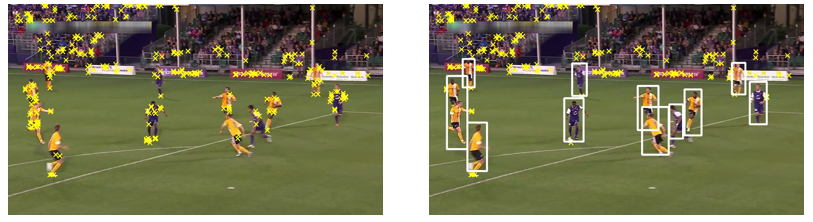}  
    \vspace{-0.1in}
	\caption{Player detection. Left: keypoints without player detection; right: keypoints with player detection. Best viewed in color.}
	\label{fig:player_detection}
    \vspace{-0.2in}
\end{figure}

We integrate player detection with the SLAM system. First, we detect objects using a pre-trained Faster R-CNN network \cite{ren2015faster} on PASCAL VOC dataset. Then, we extract bounding boxes of the \doubleQuote{person} category and remove the keypoints that are inside of the bounding boxes in tracking/mapping. As a result, we remove most of keypoints that are on player bodies while only sacrifice a few background keypoints. Figure \ref{fig:player_detection} shows an example of keypoints with/without player detection.

\section{Experiments}
We conducted experiments on one synthetic dataset and two real datasets.

{\bf Basketball dataset:}
This dataset has one image sequence (1280 $\times$ 720 at 60 fps) from a high school basketball match \cite{chen2015mimicking}. The sequence length is 3,600 frames and the pan angle range is about $[-30^o, 20^o]$. This dataset is challenging in terms of dynamic objects, fast camera motion and repeated background objects. 

{\bf Soccer dataset:}
This dataset has one image sequence (1280 $\times$ 720 at 30 fps) from a United Soccer League (USL) soccer game \cite{keyu17light}. The sequence length is 333 and the pan angle range is about $[50^o, 70^o]$. This dataset is challenging in terms of texture-less playing ground, fast camera motion and narrow field-of-view. Moreover, the focal length dramatically changes from 4,000 pixels to 2,000 pixels in about 2 seconds.  

On both datasets, the ground truth is manually annotated at each frame independently. The reprojection error is about 1.5 and 2 pixels for the basketball dataset and the soccer dataset, respectively. The PTZ base parameters are pre-estimated and fixed in the experiment.  

{\bf Error metric:}
For our method, we report the mean and standard deviation of errors for pan, tilt and focal length. We also visualize estimated camera trajectories to evaluate the smoothness of the results.

We compare our method with four baselines. For a fair comparison, player detection results are also used in all baselines except for baseline 1.

{\bf Baseline 1:} We remove the player detection component in our system and keep the remaining settings the same.

{\bf Baseline 2:}
We adapt a 3D EKF-based SLAM method \cite{davison2007monoslam} to the PTZ camera model. The method only tracks playing-ground landmarks at $Z=0$. The non-playing-ground points are not used as we cannot estimate their 3D locations from a pure-rotation camera.

{\bf Baseline 3 (EKF-H): } 
We implement an online version of the method \cite{lisanti2016continuous} which uses an off-line mapping for PTZ camera tracking. In this baseline, all the landmarks are located on the plane (the mosaic plane in \cite{lisanti2016continuous}) that is extended from the first frame. In tracking, we first use RANSAC to estimate the homography from the current frame to the mosaic plane. Then, EKF is used to update homography and landmarks based on observations. We denote this method by EKF-H because it essentially replaces our camera model by a homography.

{\bf Baseline 4 (off-line):} We implement a reference frame based method \cite{chen2015mimicking}. We manually select five reference frames that cover the whole scene. In testing, each frame finds the most-similar frame from the reference frames with SIFT feature matching. Then, the camera pose is estimated by frame-to-frame keypoint matching. This baseline is off-line as it needs a set of references with annotated camera poses before tracking.

\begin{figure}[t]
	\centering
	\includegraphics[width=0.95\linewidth]{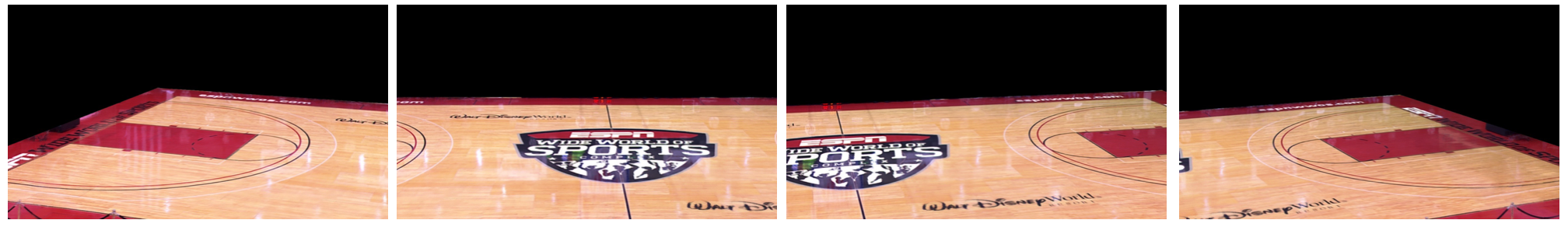}  
    \vspace{-0.1in}
	\caption{Synthetic image examples.}
	\label{fig:syn_image}
    \vspace{-0.1in}
\end{figure}

%In sequence 5 and 6, the tracking is lost in EKF-H because of the fast camera motion.
\begin{table}[t]
\parbox{.63\linewidth}{
\centering
 \scalebox{0.7}{
  \begin{tabular}{|c|c|ccc|ccc|}
    \hline
    \multicolumn{2}{|c|}{\multirow{2}{*}{Sequence}} & \multicolumn{6}{c|}{Reprojection error (pix.)}\\ \cline{3-8}
    \multicolumn2{|c|}{} & \multicolumn{3}{c|}{EKF-H} & \multicolumn{3}{c|}{EKF-PTZ (ours)} \\ \cline{1-8}
    
    Seq. ID       & Velocity & Mean & Median &  Max & Mean & Median &  Max \\ \hline
    1   & 0.02 & 0.1    & 0.1      & 0.1      & 0.1  &    0.1  & 0.2    \\ \hline
    2   & 0.83 & 0.4  & 0.4    &  0.7 & 0.3  & 0.1    &  1.1  \\ \hline
    3   & 0.70 &  1.0   & 0.3      & 16.0      & 0.3  &  0.3    & 0.5    \\ \hline
    4   & 0.08 &  2.1   &  2.2      & 3.9      & 0.7  & 0.7     & 1.3    \\ \hline
  \end{tabular}
  }
  \vspace{4mm}
  \caption{EKF-H vs. our method for camera tracking. The (mean angular) velocity (degrees per second) is measured from ground truth. }
    \label{table:homography_vs_ours}
}
\hspace{0.05in}
\parbox{.25\linewidth}{
\centering
 \scalebox{0.7} {
  \begin{tabular}{|c|c|c|c|}
    \hline %Outlier  (\%)
    \multirow{2}{*}{\begin{tabular}{@{}c@{}}Outlier \\ (\%)\end{tabular}} & \multicolumn{2}{c|}{Correctness (\%)} \\ \cline{2-3}
    
              &  Keyframe &  RF (ours) \\ \cline{1-3}
    10        &    98.0          & 100 \\
    20        &    89.3       & 100 \\
    30        &    79.0       & 100 \\
    40        &    67.7       & 100 \\
    50        &    47.0       & 99.7 \\
    \cline{1-3}
  \end{tabular}
  }
  \vspace{0.5mm}
  \caption{Keyframe-based method vs. ours for relocalization.}
  \label{table:relocalization_cmp}
}
\vspace{-0.15in}
\end{table}

\subsection{Synthetic Data Experiments}
We conducted experiments on a synthetic basketball dataset to verify the robustness of our method. Specially, we want to show that our camera model (Section \ref{subsec:camera_model}) and pan-tilt forest (Section \ref{subsec:rf}) are superior to homography-based and keyframe-based methods, respectively. First, we use multiple (about 30) calibrated images to generate a top-view image of the playing ground with detailed textures. Then, we synthesize sequences of images by warping the court image using pre-defined camera poses. Finally, we run our method on the generated images. Figure \ref{fig:syn_image} shows four image examples from the synthetic sequences.

We test the camera tracking accuracy on four sequences. Each sequence has 600 frames (10 seconds) and has various angular velocities. We run our method and baseline 3 (EKF-H) on these sequences. In testing, we turn off the relocalization of both methods. Then, we report the reprojection errors which are computed by projecting rays to images using ground truth cameras and estimated cameras. Table \ref{table:homography_vs_ours} shows the mean angular velocity (degrees per second) in each sequence and the mean, median and maximum reprojection errors (pixels). Our method achieves lower reprojection errors than the EKF-H method. It demonstrates the advantage of our camera model. Moreover, we found it is hard to tune the parameters in EKF-H (\eg the uncertainty of homography). On the other hand, it is easier to set the uncertainty of pan-tilt-zoom parameters in our method because they are physically meaningful.   

We also compare the robustness of our relocalization method with traditional keyframe-based method \cite{younes2017keyframe} using a $3,600$ frame sequence. The keyframe-based method uses the nearest keyframe to relocalize the lost frame. In the testing, the same keyframe selection method is used to make sure the same set of keyframes are used for both methods. Random outliers are added to the keypoints in both methods. Table \ref{table:relocalization_cmp} shows the percentage of correctly relocalized testing images. An estimated camera is \doubleQuote{correct} when it is within $2^{\circ}$ angular error of the ground truth. This accuracy is sufficient to restart the tracking system \cite{shotton2013scene}. Our method achieves much higher correctness than the keyframe-based method (\eg 100\% vs. 68\% when outlier ratio is 40\%) because the random forest effectively uses multiple frames in prediction.

\subsection{Real Data Experiments}

\begin{table}
 \centering
 \scalebox{0.7}{
  \begin{tabular}{|c|c|c|c|c|c|c|}
    \hline
   \multirow{2}{*}{Method} & \multicolumn{3}{c|}{Basketball} & \multicolumn{3}{c|}{Soccer} \\ \cline{2-7}
    
       & Pan ($^\circ$) & Tilt ($^\circ$) &  FL (pix.)  & Pan ($^\circ$) & Tilt ($^\circ$) &  FL (pix.) \\ \hline
    
    Baseline 1   & $0.50\pm0.46$  & $0.02\pm0.02$ & $31.95\pm25.37$ & $ 0.28\pm0.32$ &  $ 0.10\pm0.11$      &  $ 110.58\pm62.66$   \\ \hline
    
    Baseline 2   & $0.48\pm0.41$ & $0.05\pm0.03$ & $38.44\pm23.58$ & - & -       & -    \\ \hline
   
    Baseline 3   & $ 3.63\pm3.03$ & $ 0.34\pm0.27$ & $ 395.40\pm303.88$ & $ 0.74\pm0.99$ & $ 0.13\pm0.13$ & $ 179.33\pm150.45$ \\ \hline
    
    Baseline 4 (Off-line)   &  $0.10\pm0.61$ & $0.06\pm1.31$ &  $16.63\pm96.65$ & $ 0.04\pm0.05$& $ 0.02\pm0.02$& $ 36.01\pm41.46$ \\ \hline
     
    Our method   & $0.17 \pm 0.14 $  & $0.05 \pm 0.03 $  & $19.28 \pm 18.68 $ & $0.08 \pm 0.07 $ & $0.08 \pm 0.07 $ & $63.70 \pm 61.66 $ \\ \hline
  \end{tabular}
  }
  \vspace{1mm}
  \caption{Quantitative results. The metric is mean absolute errors compared with the human annotation. FL is short for focal length. Although baseline 4 is slightly better than ours, it needs an off-line initialization and has much larger maximum errors on basketball dataset (details in the text).}
    \label{table:quantitative}
\end{table}

{\bf Quantitative results:} Table \ref{table:quantitative} shows the quantitative results of our method and four baselines. On both datasets, our method achieves significantly lower estimation errors than the online baselines except for baseline 4. The result of baseline 1 shows that the player detection part improves the accuracy of our system. Baseline 2 has higher errors on the basketball dataset and is lost on the soccer dataset because many keypoints are not at the playing ground. In baseline 3,  more errors are accumulated compared with our method. Baseline 4 is slightly better than our method in terms of mean errors. However, it has much higher maximum errors on the basketball sequence. Its maximum pan error is $15.89^{\circ}$ (ours is $0.96^{\circ}$) because baseline 4 does not consider temporal information in the process, making it unsuitable for applications such as augmented reality which requires smooth camera trajectories.  

\begin{figure}[t]
	\centering
	\includegraphics[width=0.85\linewidth]{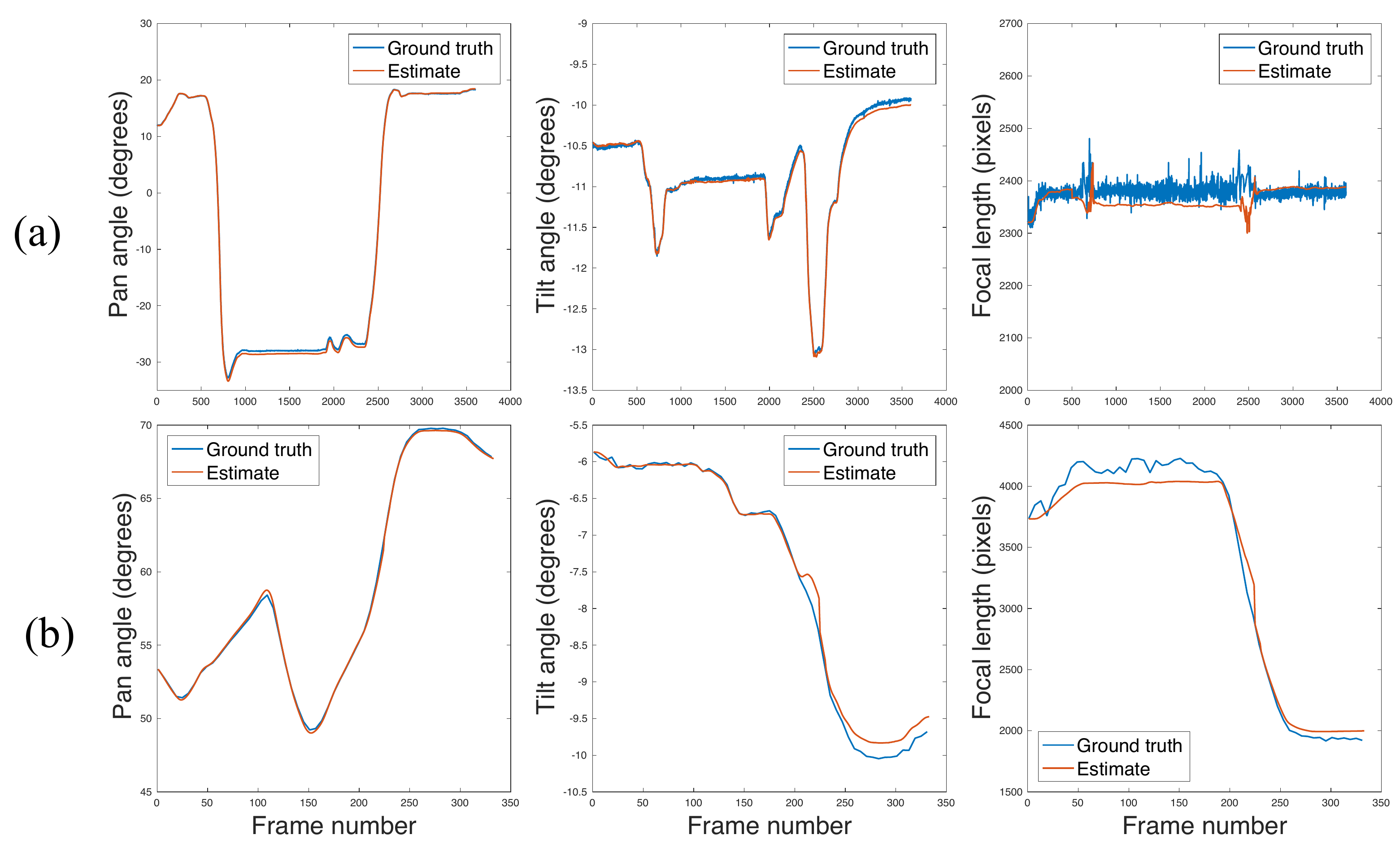}  
    \vspace{-0.1in}
	\caption{Estimated camera trajectories of our method. (a) basketball; (b) soccer.}
	\label{fig:basketball_soccer}
    \vspace{-0.15in}
\end{figure}

{\bf Qualitative results:} Figure \ref{fig:basketball_soccer} shows the estimated pan, tilt and focal length on the basketball and soccer sequences. Our result is very close to the ground truth \footnote{In the basketball sequence, the range of focal length is small (from about 2,300 pixels to 2,480 pixels). As a result, the plot of the ground truth has some noise because the ground truth camera pose is manually annotated frame by frame.}. Our result is smoother because of the EKF method. In the soccer sequence, camera rotation and the focal length change dramatically in a short time. Our method successfully tracks the camera with small errors. The results demonstrate the robustness of our system for challenging sequences in different sports.

\begin{figure}[t]
	\centering
	\includegraphics[width=0.98\linewidth]{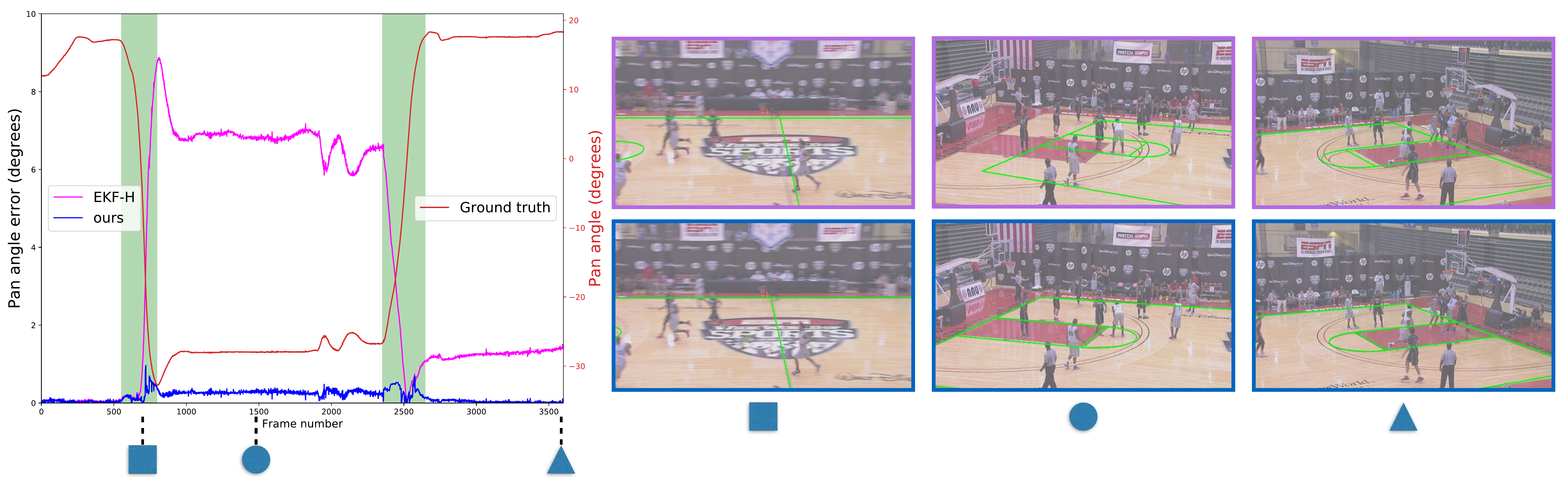}  
    \vspace{-0.1in}
	\caption{Qualitative comparison with EKF-H. The left figure shows pan angle errors of EKF-H (magenta) and ours (blue). The red line is the ground truth of pan angles that has two fast-moving periods (semi-transparent green areas). EKF-H loses the tracking at the first fast-moving period. Our method successfully tracks the camera. On the right, we visualize camera poses at three frames for both methods (EKF-H is on the top and ours is on the bottom). Best viewed in color. }
	\label{fig:qualitative_vs_ekf_h}
    \vspace{-0.15in}
\end{figure}

Figure \ref{fig:qualitative_vs_ekf_h} provides a detailed comparison of our method with EKF-H on the basketball pan angle estimation. The camera tracking of EKF-H is lost when the camera quickly moves from one side to the other side of the court (the first semi-transparent green area). After that, the estimated camera has a significant drift from the ground truth. Although EKF-H recovers the camera around frame 2,500 (the second semi-transparent green area), it still has larger errors than our method. On the other hand, our method successfully tracks the camera along the whole sequence. More results are in the supplementary material.

{\bf Implementation:} Our system is implemented with Python on a 4.20 GHz Intel CPU, 32 GB memory Ubuntu system. The player detection model is running on a TITAN Xp GPU with 12 GB memory. The running speed for our system is not optimized in the current implementation. The speed can be improved by implementing with C++ and using a GPU to compute image features. The code is available online\footnote{\url{https://github.com/lulufa390/Pan-tilt-zoom-SLAM}}.

For keypoint detection and description, we tried DoG + SIFT, FAST + ORB and LATCH \cite{levi2016latch} + ORB. We found ORB is much faster than SIFT in descriptor computation and matching. However, ORB matching is not stable when cameras rotate rapidly. On the other hand, SIFT is more reliable for rapid camera rotation. It also requires a much fewer number of keypoints to achieve similar accuracy. We finally choose DoG + SIFT in our implementation. In EKF-based tracking, the parameters are as follows: the variance of keypoints location, pan/tilt angles and focal lenth is 0.1 pixels, $0.001^\circ$ and 1 pixel, respectively.

In relocalization, we tested the nearest neighbor search (NNS) as suggested by reviewers. In training, NNS builds a feature-label database using KD-trees. In testing, the method finds pixel-ray correspondences using the nearest neighbor search in the database, followed by a RANSAC-based camera pose optimization. Although we tried ways to improve the performance, NNS is still not stable on the basketball dataset. One explanation is that the background has repeating patterns that are difficult to distinguish in the feature space. Please note, random forest methods optimize tree parameters using both features and rays. Thus, the learned model is more discriminative than KD-trees. This result agrees with the exceptional performance of random forests in camera relocalization \cite{shotton2013scene,meng2017backtracking,cavallari2019let}

{\bf Discussion:} We develop a PTZ SLAM system and successfully apply it to challenging sports videos. However, there are several limitations at the current stage. For example, our system requires the camera pose of the first frame and camera base parameters to initialize the system. Both of them can be estimated by fully automatic methods such as DSM \cite{homayounfar2017sports} and bundle adjustment \cite{triggs1999bundle}, respectively. Also, the experiments on real data are limited because there are no large public datasets available. We mitigate this limitation by careful and dense testing on a synthetic dataset.

\section{Conclusion}
In this work, we developed a robust pan-tilt-zoom SLAM system that can track fast-moving cameras in sports videos. We use a novel camera model for tracking, an online random forest for mapping and player detection to deal with moving foreground objects. The experimental results show that our method is more robust than homography-based methods. In the future, we would like to speed up our system and test it on more image sequences. Adding semantic segmentation \cite{he2017mask,kirillov2019panoptic,cioppa2019arthus} and exploring other features (\eg lines and circles) are two promising directions.

\paragraph{Acknowledgements} This work was partially supported by grants from NSERC. We thank the anonymous reviewers for insightful suggestions. 

\bibliography{egbib}
\end{document}